\pdfoutput=1

\documentclass[11pt]{article}

\usepackage[final]{acl}

\usepackage{times}
\usepackage{latexsym}
\usepackage{xcolor}
\usepackage{tcolorbox}
\usepackage{multicol}
\usepackage{url}
\usepackage{graphicx}
\usepackage{multirow}
\usepackage{amsmath}
\usepackage{subcaption}
\usepackage{capt-of} 
\usepackage{xurl}

\usepackage[T1]{fontenc}

\usepackage[utf8]{inputenc}

\usepackage{microtype}

\usepackage{inconsolata}

\usepackage{graphicx}

\title{20min-XD: A Comparable Corpus of Swiss News Articles}

\author{
 \textbf{Michelle Wastl\textsuperscript{1}}\hspace{0.3cm}
 \textbf{Jannis Vamvas\textsuperscript{1}}\hspace{0.3cm}
 \textbf{Selena Calleri\textsuperscript{2}}\hspace{0.3cm}
 \textbf{Rico Sennrich\textsuperscript{1}}\hspace{0.3cm}
\\
 \textsuperscript{1}Department of Computational Linguistics, University of Zurich\hspace{0.3cm}
 \textsuperscript{2}20 Minuten (TX Group)
\\
 \small{
 \texttt{\{wastl,vamvas,sennrich\}@cl.uzh.ch, \{selena.calleri\}@20minuten.ch}
 }
}

\begin{document}
\maketitle
\begin{abstract}

We present \textit{20min-XD} (\textbf{20 Min}uten \textbf{cross}-lingual \textbf{d}ocument-level), a French-German, document-level comparable corpus of news articles, sourced from the Swiss online news outlet \textit{20 Minuten}/\textit{20 minutes}. Our dataset comprises around 15,000 article pairs spanning 2015 to 2024, automatically aligned based on semantic similarity. We detail the data collection process and alignment methodology. Furthermore, we provide a qualitative and quantitative analysis of the corpus. The resulting dataset exhibits a broad spectrum of cross-lingual similarity, ranging from near-translations to loosely related articles, making it valuable for various NLP applications and broad linguistically motivated studies. We publicly release the dataset in document- and sentence-aligned versions and code for the described experiments\footnote{Dataset:\hspace{0.1cm}\url{https://huggingface.co/datasets/ZurichNLP/20min-XD}}$^,$\footnote{Code: \url{https://github.com/ZurichNLP/20min-XD}}.
\end{abstract}

\section{Introduction}

Cross-lingual datasets play a crucial role in Natural Language Processing (NLP), supporting a range of tasks such as bitext mining, machine translation, and cross-lingual information retrieval. Among these, comparable corpora—datasets containing text pairs with related but non-identical content across languages—are particularly valuable. Unlike parallel corpora, which consist of direct translations, comparable corpora naturally contain a mix of exact translations, paraphrases, and loosely related content, reflecting the linguistic and cultural variations between languages. This makes them a rich resource for training and evaluating multilingual NLP models~\cite{lewis2020pretrainingparaphrasing, liu-etal-2020-multilingual-denoising, philippy-etal-2025-luxembedder}.

However, existing document-level, cross-lingual corpora remain limited in scope. Many available resources are English-centric, primarily covering English alongside another high-resource language and/or are restricted to sentence-level alignments rather than full documents~\cite{zweigenbaum-etal-2017-overview, artetxe-schwenk-2019-massively}. At the same time, large language models (LLMs) and modernized encoder architectures are advancing in their ability to process longer texts and numerous languages, further increasing the demand for multi-/cross-lingual, document-level corpora~\cite{hengle2024multilingualneedlehaystackinvestigating, wang-etal-2024-improving-text, zhang-etal-2024-mgte}. 

Beyond their NLP applications, cross-lingual document-level datasets also facilitate more linguistically motivated studies such as cross-cultural discourse analyses~\cite{carbaugh} or comparative journalism research~\cite{ComparativeJournalismResearch}. More specifically, a German-French news article corpus could be used to examine how news narratives and framing strategies vary between the Germanophone and Francophone regions.

Given these potential interdisciplinary use cases, we collect comparable news articles in German and French from the online Swiss news outlet \textit{20 Minuten}/\textit{20 minutes}. As both editions are produced by the same publisher, with an internal article transfer workflow from one language to the other, they share a high degree of topical overlap, making them well-suited for comparable corpus creation. Our dataset comprises 15,000 article pairs, spanning nearly a decade (2015–2024). Each article pair consists of a German and a French news article published on the same day, covering the same or a highly related event. In addition to the document-level alignments, we release a sentence-aligned version of the dataset, which contains 117,126 sentences per language.

We release the dataset to the research community for non-commercial, scientific purposes\footnote{See Appendix~\ref{app:copyright} for the detailed Copyright notice.}.
\begin{table*}[t!] 
    \renewcommand{\arraystretch}{1.1} 
    \centering
    \LARGE
    \resizebox{\linewidth}{!}{
    \begin{tabular}{@{}lcccccc@{}}
    \hline
    & \multicolumn{2}{c}{\phantom{000}\textbf{Validation Set}} & \multicolumn{2}{c}{\phantom{0000}\textbf{Full Dataset}} & \multicolumn{2}{c}{\phantom{0000}\textbf{Top 15k}} \\  
    \textbf{Statistics} & \phantom{000}\textbf{German} \phantom{0,} & \phantom{000}\textbf{French} \phantom{,} & \phantom{00} \textbf{German} & \phantom{00,,} \textbf{French} & \phantom{000} \textbf{German} & \phantom{000} \textbf{French} \\  
    \hline
    Total \# of aligned articles & \phantom{000,}14 &  \phantom{000}14 & \phantom{000,0}73,085 & \phantom{000,0}73,085 & \phantom{00,0}15,000 & \phantom{000,}15,000 \\
    Total \# of sentences & \phantom{00,}401 & \phantom{00,}358 & \phantom{00,}1,888,323 & \phantom{00,}1,608,497 & \phantom{00,}357,071 & \phantom{00,}327,628 \\
    Total \# of tokens & \phantom{0}9,087 & \phantom{0}9,690 & \phantom{0}43,559,153 & \phantom{0}43,256,366 & \phantom{0}8,378,874 & \phantom{0}8,956,116 \\
    Total \# of characters & 38,523 & 38,519 & 189,598,932 & 174,789,207 & 36,924,383 & 36,387,070 \\
    Avg. title length in characters & \phantom{000,}59 & \phantom{000,}54 & \phantom{000,000,0}51 & \phantom{000,000,0}53 & \phantom{000,000,}51 & \phantom{000,000,}54 \\
    Avg. title length in tokens & \phantom{000,}18 & \phantom{000,}18 & \phantom{000,000,0}15 & \phantom{000,000,0}17 & \phantom{000,000,}15 & \phantom{000,000,}17 \\
    Avg. lead length in characters & \phantom{00,}146 & \phantom{00,}155 & \phantom{000,000,}152 & \phantom{000,000,}146 & \phantom{00,000,}152 & \phantom{00,000,}150 \\
    Avg. lead length in tokens & \phantom{000,}39 & \phantom{000,}43 & \phantom{000,000,0}39 & \phantom{000,000,0}40 & \phantom{00,000,0}38 & \phantom{00,000,0}41 \\
    Avg. content length in characters & \phantom{0}2,547 & \phantom{0}2,542 & \phantom{0000,0}2,391 & \phantom{0000,0}2,192 & \phantom{0000,}2,258 & \phantom{0000,}2,222 \\
    Avg. content length in tokens & \phantom{00,}706 & \phantom{00,}753 & \phantom{000,000,}650 & \phantom{000,000,}649 & \phantom{00,000,}612 & \phantom{00,000,}655 \\
    Avg. content length in sentences & \phantom{000,}29 & \phantom{000,}26 & \phantom{000,000,0}26 & \phantom{000,000,0}22 & \phantom{000,000,}24 & \phantom{000,000,}22 \\
    \hline
    \end{tabular}
    }
    \caption{\label{tab:stats} Detailed statistics of the validation, full, and top-15k subsets. The sentence segmentation was performed with spaCy '[de/fr]\_core\_news\_sm'~\cite{spacy2} models for sentence segmentation and tokenization with the \texttt{paraphrase-multilingual-mpnet} tokenizer.}
\end{table*}

\section{Related Work}

Switzerland’s multilingual landscape, with four official languages, provides fertile ground for cross-lingual corpus creation. Several prior works have leveraged this linguistic diversity to construct multilingual datasets. For instance, SwissAdmin~\cite{Scherrer2014} is a sentence-aligned corpus of official Swiss government press releases available in German, French, Italian, and English. Similarly, the Bulletin Corpus~\cite{volk2016building} aligns issues of the \textit{Credit Suisse Bulletin} across the same four languages.\\
\indent\textit{20 Minuten} has also served as a resource for previous NLP-related studies. \citet{rios-etal-2021-new} constructed a dataset for automatic text simplification by pairing original German \textit{20 Minuten} articles with their simplified counterparts. More recently, \citet{kew-etal-2023-20} created a dataset aimed at automatic news summarization in German, further expanding the utility of Swiss news data in NLP research.\\
\indent With this work, we aim to bridge these two subjects by introducing \textit{20min-XD}, a French-German document-level comparable corpus, sourced from \textit{20 Minuten} (German) and \textit{20 minutes} (French). 

\section{Data Acquisition}
To construct our dataset, we first scrape a total of 593,897 online news articles from both \url{www.20min.ch/} and \url{www.20min.ch/fr/}, covering the period from 01.01.2015 to 01.12.2024.  In the following subsections, we describe the process applied to identify and align the semantically related articles.

\subsection{Validation Set}
To establish a gold standard for alignment evaluation, we selected all articles from a single publication day, resulting in 87 German and 70 French articles. Each French article was manually compared against the German articles to identify comparable pairs. While we did not strictly prohibit n:n pairings, the resulting validation set only contains 1:1 pairings. Through this process, we aligned 28 articles into 14 pairs, forming our validation set. Detailed statistics can be found in Table~\ref{tab:stats}.

\begin{figure*}[t!]  
    \centering
    \includegraphics[width=1\linewidth]{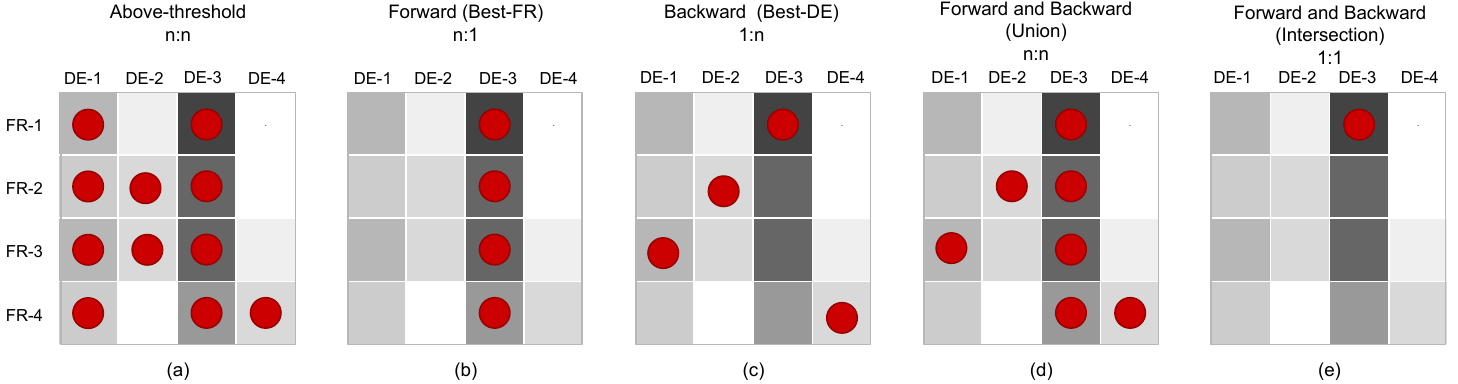} 
    \caption{Matrix visualization of different alignment strategies.}
    \label{fig:alignment-vis}
\end{figure*}

\begin{table*}[t!] 
    \LARGE
    \renewcommand{\arraystretch}{1.1} 
    \centering
    \resizebox{0.9\linewidth}{!}{
    \begin{tabular}{@{}lccccc|c@{}}
    \hline
    \textbf{Model} & \textbf{Above-threshold} & \textbf{Intersection} & \textbf{Union} & \textbf{Best-DE} & \textbf{Best-FR} & \textbf{Avg.} \\
    \hline
    \texttt{paraphrase-multilingual-mpnet} & 54.1 & \textbf{64.7} & 54.1 & 57.8 & 55.8 & 57.3 \\
    \texttt{gte-multilingual-base} & 55.6 & 62.1 & 55.6 & 60.0 & 58.5 & 58.3 \\
    \texttt{LaBSE} & 53.3 & 48.5 & 56.5 & 60.0 & 46.2 & 52.9 \\
    \texttt{sentence-swissBERT} & 62.9 & 62.5 & 62.9 & 61.1 & 62.5 & 62.4 \\
    \texttt{gte-modernbert-base} & 45.5 & 53.3 & 50.0 & 54.1 & 50.0 & 50.6 \\
    \hline
    \end{tabular}
    }
    \caption{\label{tab:model_comparison} F1 performance comparison of different models and different alignment approaches on the validation set. The corresponding thresholds are presented in Appendix~\ref{app:thresholds}.}
\end{table*}

\subsection{Automatic Article Alignment}
\label{subsec:align}
Since manually aligning comparable articles across languages is time-intensive and requires proficiency in both German and French, we automate the process leveraging multilingual embedding models. Specifically, we encode portions of each article as numerical vectors and compute cosine similarity scores, which range from -1 to 1 ($*100$), to quantify their semantic similarity. 

In order to find the most appropriate alignment methods for the \textit{20 Minuten} articles, we conduct experiments on our validation set with different embedding models, alignment approaches, and similarity thresholds.

We choose not to embed the full article texts to ensure a fair comparison across the tested models, some of which have a sequence length constraint (3 out of the tested 5). The results on our validation set suggest that concatenating the article's title and lead provides a sufficiently strong signal for document alignment. This enables resource-efficient experimentation with encoder-based embedding models while avoiding length limitations.

\vspace{0.2cm}

\subsubsection{Models}

We experiment with the set of models presented in Table~\ref{tab:model_comparison}: \texttt{paraphrase-multilingual-mpnet} is a state-of-the-art multilingual sentence-level paraphrase recognition model~\cite{song2020mpnetmaskedpermutedpretraining}; 
\texttt{gte-multilingual-base}, a long-context multilingual text representation model~\cite{zhang-etal-2024-mgte};
\texttt{sentence-swissBERT}, a sentenceBERT-based~\cite{reimers-2019-sentence-bert} model trained on in-domain (\textit{20 Minuten}) data~\cite{grosjean-vamvas-2024-fine};
\texttt{gte-modernbert-base}, modernized, more efficient, long-context version of BERT that has been trained on predominantly English data~\cite{warner2024smarterbetterfasterlonger}.

\vspace{0.2cm}

Preliminary experiments with an LLM-based model~\cite{wang-etal-2024-improving-text} have shown that they outperform encoder-based models while also being able to process longer input sequences. They do, however, also increase the computational complexity of the embedding process, making it rather resource-intensive and barely feasible in terms of memory and time if scaled to a larger number of documents.

\subsubsection{Alignment Strategies}

Previous work in cross-lingual alignment has considered multiple possible alignment strategies that either expand or restrict the resulting number of alignments according to different categories as described in e.g.~\citet{jalili-sabet-etal-2020-simalign} for cross-lingual word alignment. Similarly to ~\citet{hammerl-etal-2024-understanding}, we experiment with strategies that result in a range from weak to strong alignment, where strategies for weaker alignments typically allow a higher range of semantic similarity and multiple possible alignments, while strategies for stronger alignments are more restrictive towards a high semantic similarity and may only include one good alignment (Figure~\ref{fig:alignment-vis}). 

\vspace{0.2cm}

\noindent\textbf{Above-Threshold} considers all document pairs with a similarity score above a certain threshold as alignable, allowing for many-to-many (n:n) alignments. This means that any number of French articles can be linked to any number of German articles without additional constraints beyond the similarity threshold. While this approach captures a broad range of potential alignments, it does not enforce uniqueness or best-match constraints, leading to a higher number of alignments (Figure~\ref{fig:alignment-vis}a).

\vspace{0.2cm}

\noindent\textbf{Best-FR} applies a many-to-one (n:1, German:French) constraint, where each FR article is aligned to the single DE article with which it has the highest cosine similarity, provided that the similarity exceeds the threshold. This ensures that each FR document has a single best-matching DE counterpart, but multiple French articles can still be mapped to the same German article. This approach prioritizes French articles selecting their closest German equivalent while allowing asymmetry in alignments (Figure~\ref{fig:alignment-vis}b).

\vspace{0.2cm}

\noindent\textbf{Best-DE} follows the same principle as Best-FR but from the German perspective, enforcing a one-to-many (1:n, German:French) constraint. This results in a setting where a single German article may be linked to multiple French articles, capturing scenarios where a single German document is the best translation candidate for multiple French counterparts (Figure~\ref{fig:alignment-vis}c).

\vspace{0.2cm}

\noindent\textbf{Union} takes the union of Best-DE and Best-FR alignments, allowing many-to-many (n:n) alignments, but in a more restrictive manner than the Above-Threshold approach. Instead of considering all pairs above the threshold, it only retains document pairs where at least one side selects the other as its most similar document above the threshold (Figure~\ref{fig:alignment-vis}d). 

\vspace{0.2cm}

\noindent\textbf{Intersection} is the most restrictive strategy, enforcing a one-to-one (1:1) constraint. A valid alignment occurs only when the French article is the best match for the German article and vice versa provided their similarity score exceeds the threshold. This method forms the intersection of Best-DE and Best-FR, ensuring that alignments are bidirectional and mutually optimal (Figure~\ref{fig:alignment-vis}e).

\subsection{Setting a threshold}
Since not every article has a comparable counterpart in the other language, we define a similarity score threshold above which two articles are considered alignable. This threshold must be exceeded in each of the alignment strategies described above. To determine the optimal threshold $\theta$, we iterate through the range of 0 and 100 in steps of 0.5, selecting the one that maximizes the F1 score on our validation set:

\[
\hat{\theta} = {\arg\max}_{\theta\in\{0, 0.5,\dots, 100\}} F_1(\theta)
\]

\vspace{0.2cm}

And we define F1 as follows, where $P$ denotes predicted pairs and $G$ gold pairs:

\[
Prec = \frac{|P\cap G|}{|P|}
\]

\[
Recall = \frac{|P\cap G|}{|G|}
\]

\[
F_1 = 2 \cdot \frac{Prec \cdot Recall}{Prec + Recall}
\]

\vspace{0.2cm}

This process is repeated for each of the embedding models described above. Our results show that \texttt{paraphrase-multilingual-mpnet} with the alignment strategy \textit{intersection} at a similarity score threshold of 46, outperforms all other models on the validation set (see Table~\ref{tab:model_comparison}), making it our approach for article alignment.
 
It is worth noting that the number of samples in our validation set is small (87 German and 70 French articles). This could lead to statistical noise, exaggerating the apparent differences in the results, making them seem larger/smaller than they truly are.

\subsection{Choosing A Time Window}
To ensure precise alignment and reduce computational complexity, we restrict comparisons to articles published on the same date. This approach minimizes spurious matches between articles that discuss similar topics but are unrelated in terms of specific events or developments.

\subsection{Post-Processing}
After aligning the French and German articles, we clean the resulting corpus. Manual inspection indicates that faulty articles usually have a suspiciously high similarity score and contain an error message or the same text in the same language. We remove such pairs.

\subsection{Sentence Alignment}
To provide more fine-grained insights into the dataset, we conduct sentence-level analyses. To achieve this, we first segment articles into sentences using the spaCy ‘[de/fr]\_core\_news\_sm’~\cite{spacy2} models for German and French.

Once segmented, we perform cross-lingual sentence alignment, once again, applying the best performing approach described above: \texttt{paraphrase-multilingual-mpnet} with the \textit{intersection} alignment strategy. While we consider only sentence pairs with a similarity score above 46 for our analyses, we release the sentence-aligned version of our corpus on all aligned sentences, including those whose similarity score does not exceed the threshold. This allows for more holistic future analyses, capturing not only the most strongly aligned sentences but also those with the weakest still detectable semantic similarity.

We post-process the sentence-level version of the dataset by removing sentence pairs that contain less than 30 characters, which entails names, trailing characters and source abbreviations.

\begin{table*}[t!] 
    \LARGE
    \renewcommand{\arraystretch}{1.2} 
    \resizebox{\linewidth}{!}{
    \begin{tabular}{@{}l p{12cm} p{12cm} @{}}
    \hline
    \rule{0pt}{1.5ex} 
    \textbf{Similarity Scores} & \textbf{German} & \textbf{French} \\
    \hline
    \rule{0pt}{1.5ex} 
    \begin{minipage}[t]{10cm}
        \raggedright
        Cosine: 98.48 (\textbf{max})\\
        SentLengthCorr: 0.75\\
        AlignRatio DE: 0.68\\
        AlignRatio FR: 0.56\\
        Monotonicity: 1.0
    \end{minipage}& 
    \begin{minipage}[t]{12cm} 
        \raggedright
        \textbf{Title:} Mobilität.: «Ab 2030 bieten wir nur noch vollelektrische Fahrzeuge an» \\  
        \textbf{Lead:} Die Elektro-Revolution rollt. Traditionelle Autohersteller haben derzeit einen schweren Stand. Wir haben bei Helen Hu, Geschäftsführerin des Schweizer Ablegers von Volvo, seit 2010 in chinesischer Hand, nachgefragt, wie sie die Zukunft der Mobilität sieht.
    \end{minipage} &
    \begin{minipage}[t]{12cm} 
        \raggedright
        \textbf{Title:} Mobilité: «A partir de 2030, nous ne proposerons plus que des véhicules entièrement électriques» \\ 
        \textbf{Lead:} La révolution électrique est en marche. Les constructeurs automobiles traditionnels ont actuellement la vie dure. Nous avons demandé à Helen Hu, directrice de la filiale suisse de Volvo, en mains chinoises depuis 2010, comment elle voit l’avenir de la mobilité.
    \end{minipage} \\
    \noalign{\vskip 0.8ex}
    \hline
    \rule{0pt}{1.5ex} 
    \begin{minipage}[t]{10cm}
        \raggedright
        Cosine: 84.05 (\textbf{mean among top-15k})\\
        SentLengthCorr: -0.78\\
        AlignRatio DE: 0.23\\
        AlignRatio FR: 0.21\\
        Monotonicity: -1.0
    \end{minipage}& 
    \begin{minipage}[t]{12cm} 
        \raggedright
        \textbf{Title:} LKW kreuzte Lieferwagen und stürzte dann ab \\  
        \textbf{Lead:} Ein Lastwagen stürzte am Dienstag 300 Meter in die Tiefe. Der 66-jährige Fahrer wurde schwer verletzt. Jetzt gibt es erste Erkenntnisse, wie es zum Unfall kam.
    \end{minipage} &
    \begin{minipage}[t]{12cm} 
        \raggedright
        \textbf{Title:} Un camion chute de 300 mètres, le chauffeur survit\\ 
        \textbf{Lead:} Un chauffeur de poids lourd a été grièvement blessé, mardi, après que son véhicule est sorti de la route, dans le canton d'Uri.
    \end{minipage} \\
    \noalign{\vskip 0.8ex}
    \hline
    \rule{0pt}{1.5ex} 
    \begin{minipage}[t]{10cm}
        \raggedright
        Cosine: 78.65 (\textbf{min among top-15k})\\
        SentLengthCorr: -0.47\\
        AlignRatio DE: 0.07\\
        AlignRatio FR: 0.2\\
        Monotonicity: -0.3
    \end{minipage}& 
    \begin{minipage}[t]{12cm} 
        \raggedright
        \textbf{Title:} GP Brasilien - Bottas gewinnt das Sprintrennen – Hamilton nach irrer Aufholjagd auf Rang 5 \\ 
        \textbf{Lead:} Am Samstag stand beim GP von Brasilien die Sprint-Entscheidung an. Die 3 WM-Punkte und die Pole-Position für das Rennen am Sonntag sicherte sich Valtteri Bottas.
    \end{minipage} &
    \begin{minipage}[t]{12cm} 
        \raggedright
        \textbf{Title:} Automobile – Bottas prive Verstappen de la victoire au sprint et de la pole \\
        \textbf{Lead:} Valtteri Bottas s’est offert la course sprint et partira de la première case dimanche au Grand Prix du Brésil. Max Verstappen sera placé derrière lui et Lewis Hamilton 10e.
    \vspace{0.1cm}
    \end{minipage} \\
    \noalign{\vskip 0.8ex}
    \hline
    \hline
    \rule{0pt}{1.5ex} 
    Cosine: 46.00 (\textbf{min among full dataset}) & 
    \begin{minipage}[t]{12cm} 
        \raggedright
        \textbf{Title:} Sein Zwillingsbruder brachte ihn vor Gericht \\ 
        \textbf{Lead:} Hochriskante Börsengeschäfte ihres Verwaltungsratspräsidenten haben eine renommierte Churer Treuhandfirma in den Ruin getrieben. Der Beschuldigte musste vor Gericht erscheinen.
    \end{minipage} &
    \begin{minipage}[t]{12cm} 
        \raggedright
        \textbf{Title:} Plombé par Kairos, Julius Bär doit se rattraper \\ 
        \textbf{Lead:} La filiale italienne de Julius Bär apparaît presque comme la source de tous les maux du gestionnaire de fortune zurichois.
    \end{minipage} \\
    \noalign{\vskip 0.8ex}
    \hline
  \end{tabular}
  }
  \caption{\label{tab:quali} Comparison of the title and lead text of the aligned articles receiving the lowest, mean and highest cosine similarity scores from the top 15,000 aligned articles as well as the aligned articles with the lowest overall score from the full set of aligned articles, which is filtered from the final dataset.}
\end{table*}

\subsection{Additional Measures of Similarity}
\label{subsec:sim-metrics}

In the corpus description in Section~\ref{sec:results} we make use of additional cross-lingual similarity measures apart from the cosine distance that are based on the sentence alignments:

\vspace{0.2cm}

\noindent\textbf{Alignable sentences per document} To estimate how much text within an article is highly similar, we compute the relative percentage of alignable sentences. This measure is particularly interesting, as the full document is not considered during automatic article alignment, as described in Subsection~\ref{subsec:align}. For each article, we define the alignable sentence ratio as:

\[
\text{AlignRatio} = \frac{\text{NumAlignedSentences}}{\text{TotalSentences}}
\]

\vspace{0.2cm}

\noindent\textbf{Sentence length correlation} If the sentence length, measured as the number of characters in the sentence,  differs between the two languages in a systematic way, a high correlation between sentence lengths in aligned articles could be an additional indicator of semantic similarity. Hence, we compute the sentence length correlation of an article as a Pearson correlation.



\vspace{0.2cm}



\noindent\textbf{Monotonicity} We measure the cross-lingual monotonicity (degree by which aligned sentences appear in the same order) between an aligned article pair by calculating the Kendall rank correlation of the aligned sentences' position. 



\section{Dataset}
\label{sec:results}

Our alignment process results in 74,507 article pairs. During post-processing the corpus is filtered down to 73,085 article pairs.
By agreement with \textit{20 Minuten}, our dataset release is limited to 30,000 articles. Consequently, we select the top 15,000 article pairs sorted by their similarity score for publication, which we refer to as top 15k dataset in the following. 
Nonetheless, in the remainder of this paper, we will consider both the full dataset and the top 15k article pairs as subject of analysis. The detailed dataset statistics for both are presented in Table~\ref{tab:stats}. 

Out of the total 300,000+ sentences in each language from the top 15k dataset, we align 133,693 sentences per language, from which 117,126 are left after filtering. For the correlation studies in Section~\ref{subsec:corr}, we consider all the sentence pairs with similarity score above 46, totaling to 109,871 sentence pairs.

\subsection{Qualitative Analysis}

Table~\ref{tab:quali} provides a qualitative comparison of article pairs with the lowest, mean, and highest cosine similarity scores in the top 15k dataset as well as the article pair with the lowest similarity score of all 75,085 initially aligned articles. The highest-scoring pairs exhibit strong lexical and syntactic similarities. The mean-scoring pairs effectively convey the same meaning but demonstrate noticeable differences of the order in which the information is presented. Only the last sentence in the German lead as well as the last phrase in the French lead introduce different information. The lowest-scoring pair in the top 15k dataset covers the same event but differs strongly in word choice and the order in which the information is conveyed. The lowest-scoring pair of the full set of aligned articles, while still loosely related (financial crises), differs in the actual event that is described (e.g., court case leading to a company's collapse vs. corporate struggle with subsidiary).

These results suggest that our dataset mostly consists of articles covering the same topic but with varying degrees of semantic overlap, text structure and length. In order to gain further insight into these features and their relationship to semantic similarity, we conduct a correlation study between the cosine scores of the aligned articles and the different measures described in Section~\ref{subsec:sim-metrics}. 

\begin{figure}[t!]  
    \centering
    \includegraphics[width=1\linewidth]{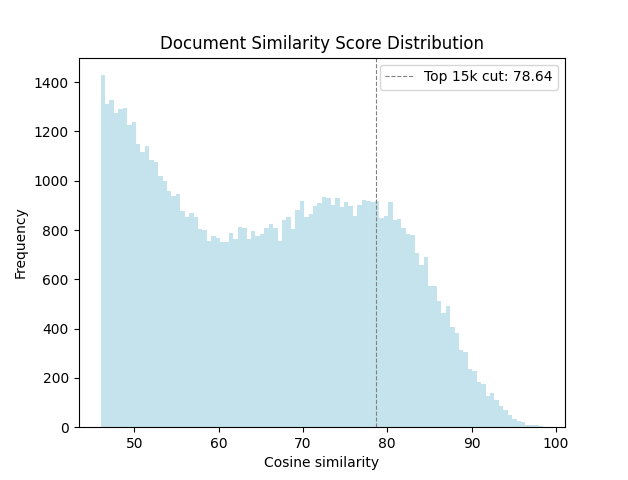} 
    \caption{Document (cosine) similarity score distribution over all 74,085 article pairs divided into 100 bins ranging from the threshold of 46 to 100. The dashed line indicates the cut above which the top 15,000 article pairs form the final comparable dataset.}
    \label{fig:dist}
\end{figure}

\begin{figure}[t!]  
    \centering
    \includegraphics[width=1.01\linewidth]{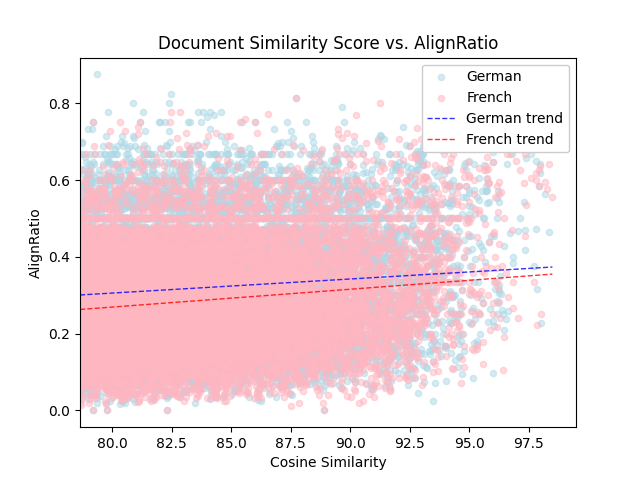} 
    \caption{The document cosine similarities in comparison to the AlignRatio of each aligned article in German and French. Both languages show a positive trend line with weak positive correlation (FR: Pearson correlation coefficient $r = 0.145$; DE: $r = 0.103$). }
    \label{fig:doc-align}
\end{figure}

\subsection{Quantitative Analysis}
\label{subsec:corr}

\subsubsection{Cosine Similarity Distribution}
Figure~\ref{fig:dist} presents the distribution of cosine similarity scores among the aligned articles. The distribution exhibits a right-skewed pattern, suggesting that among the collection of scraped articles, French and German articles with moderate semantic relatedness are more prevalent than those with extremely high similarity scores. The number of articles first drops and then rises again with a rising cosine similarity before reaching a small peak at around 80, located almost exactly at our top 15,000 cutoff point. Following the cutoff, the frequency of article pairs declines sharply to a relatively low level towards higher similarity scores. This pattern loosely suggests the presence of two clusters of article pairs: one representing moderately related articles and another, less prominent, group of more closely related articles. 

\subsubsection{Correlation with AlignRatio}

As a further measure of semantic similarity, we employ the alignment ratio (AlignRatio), which measures the proportionality of aligned sentences between the articles in the two languages, and examine how document similarity scores correlate. As shown in Figure~\ref{fig:doc-align}, both German and French exhibit weak positive correlations between cosine similarity scores and AlignRatio ($r = 0.145$  for French,  $r = 0.103$  for German). These findings suggest that articles with more alignments in the full text tend to have slightly higher semantic similarity. This supports our assumption that relying solely on the title and lead for the automatic alignment is sufficient but not perfect.

\subsubsection{Correlation with Sentence Length}

\begin{figure}[t!]  
    \centering
    \includegraphics[width=1.01\linewidth]{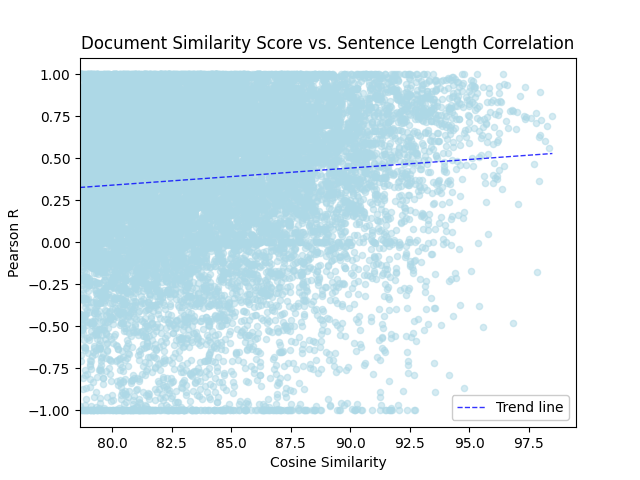} 
    \caption{The document cosine similarities in comparison to the sentence length correlation of each aligned article. There is a very weak positive trend of correlation detectable between the two variables (Pearson correlation coefficient $r = 0.084$). }
    \label{fig:doc-sent-length}
\end{figure}

To analyze the relationship between document similarity scores and sentence length variations in aligned articles, we compute the correlation between cosine similarity scores and the sentence length correlation of each article pair. As illustrated in Figure~\ref{fig:doc-sent-length}, the results indicate a very weak positive correlation ($r = 0.084$).

\subsubsection{Correlation with Monotonicity}

\begin{figure}[t!]  
    \centering
    \includegraphics[width=1.01\linewidth]{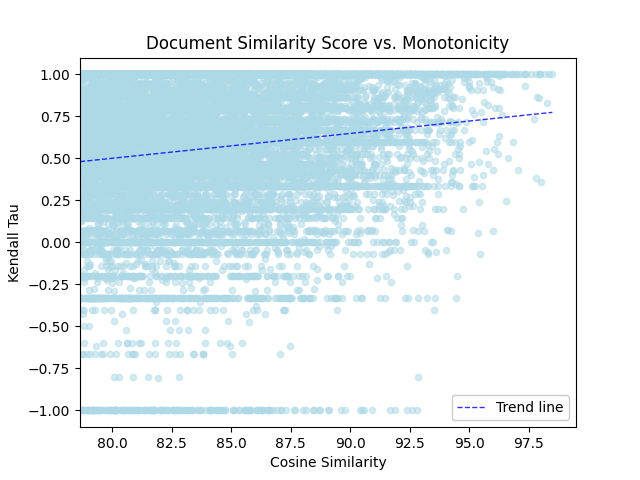} 
    \caption{The document cosine similarities in comparison to the monotonicity score of each aligned article. A weak positive correlation trend is detectable between the two variables (Pearson correlation coefficient $r = 0.147$). }
    \label{fig:doc-mono}
\end{figure}

We also investigate the relationship between document similarity scores and monotonicity, which quantifies the extent to which the order of information (= sentences) is preserved between aligned articles. Figure~\ref{fig:doc-mono}, presents the correlation between cosine similarity scores and monotonicity, showing a weak positive correlation ($r = 0.147$). This suggests, similarly to the previous results, that while higher document similarity scores are slightly associated with a more monotonic alignment of information, the effect is not strong. The clusters near -1.00 and 1.00 may indicate a high number of articles with only one or two aligned sentences — a pattern that could be worth to investigate further.

\vspace{0.2cm}

    

    



\vspace{0.2cm}

Given our qualitative analysis and correlation studies, we are confident our dataset maintains an adequate quality for a comparable corpus, covering the full range between direct translations and fairly unrelated text sequences. 
However, further work with these metrics could provide more insight. Specifically, the alignment ratio may serve as an indicator on which pieces of information are considered essential in both linguistic regions and which are missing from one or the other. Similarly, sentence length correlation could offer valuable perspectives in news-specific translation research. Lastly, monotonicity could be explored further by analyzing topic-specific trends, potentially revealing which topics tend to be translated in a more monotonic fashion than others.

\section{Future Work}
\subsection{Comparing similarity of full text}
While  using only titles and leads was sufficient for aligning comparable articles, incorporating full article content into the similarity score calculation could provide a more granular and accurate insight into the degree of semantic similarity and relatedness of the articles. This approach could provide a more nuanced representation of narrative structure, argumentation, and topical emphasis. Although computationally intensive, modern embedding models such as \texttt{e5-instruct-7b} or \texttt{gte-multilingual-base} can theoretically process longer text spans, making full-text comparison increasingly feasible.

\subsection{Multilingual long-context embedding models}
Encoder-based embedding models are currently going through a renaissance with modernized implementations, such as ModernBERT~\cite{warner2024smarterbetterfasterlonger}, with significantly improved efficiency and ability to process longer text sequences. At this point in time, multilingual versions of this model specified for the text similarity task are scarce. Future work could explore extending ModernBERT to a multilingual setting and/or optimization for cross-lingual document alignment. Another potential direction is leveraging these modern architectures to develop a document-level counterpart to the (sentence-)swissBERT model.

\subsection{Difference recognition}
While semantic similarity has been a predominant focus in NLP, the ability to detect and quantify differences between texts—especially across languages—is an emerging research area~\cite{vamvas-sennrich-2023-towards}. Inspired by diff-based operations in version control, this task could have implications for natural language versioning, collaborative document editing, and editorial workflows. \citeauthor{vamvas-sennrich-2023-towards} show that semantic similarity datasets can be repurposed for difference detection, but have to be synthetically altered to cover cross-linguality and longer text sequences. \\
Given the variation spectrum observed in our dataset (see Section~\ref{sec:results}), the diversity of near-translations and loosely related articles, an extension of our corpus with fine-grained annotations—at the paragraph, sentence, or even token level—could enable research into automatic cross-lingual difference recognition.

\section{Conclusion}

We introduce \textit{20min-XD}, a new French-German document-level comparable dataset of news articles, sourced from the Swiss newspaper \textit{20 Minuten}/\textit{20 minutes}. The dataset consists of 15,000 aligned articles (or 117,126 aligned sentences) published over a ten-year period. To establish document-level and sentence-level alignment, we employ a multilingual paraphrase recognition model, which demonstrated strong performance during experiments on a manually curated validation set. Both qualitative and quantitative results show that our corpus captures a broad spectrum of cross-lingual similarity, from near-translations to more loosely related text pairs that still cover the same event, with varying degrees of alignable sentences, text lengths and monotonicity. We anticipate its use in future studies across a broad range of linguistically motivated studies.

\section*{Acknowledgments}

This work was funded by the Swiss National Science Foundation (project InvestigaDiff; no.~10000503 for MW, JV, and RS, and project
MUTAMUR; no.~213976 for RS). We sincerely thank everyone at 20 Minuten (TX Group) for their support and for making their data accessible to the research community, with special appreciation to Dean Cavelti for his patient communication. We are also grateful to Unitectra, particularly Peter Loch, for their valuable legal guidance. Finally, we extend our thanks to the Department of Computational Linguistics at the University of Zurich for their inspiring discussions and guidance, with special recognition to Sarah Ebling, Andrianos Michail, Patrick Haller and Anastassia Shaitarova.

\bibliography{custom}

\appendix
\onecolumn

\section{Copyright Notice}
\label{app:copyright}
\definecolor{copyrightgray}{gray}{0.97} 
\definecolor{bordergray}{gray}{0.7} 

The resulting dataset is released with the following copyright notice:

\begin{quote}
    {\textbf{German / Deutsch (original):}} \\[0.3em]
    {\small © 2025. TX Group AG / 20 Minuten.}\\[0.3em]
    {\small Dieser Datensatz enthält urheberrechtlich geschütztes Material von TX Group AG / 20 Minuten. Er wird ausschliesslich für nicht-kommerzielle wissenschaftliche Forschungszwecke bereitgestellt. Jegliche kommerzielle Nutzung, Vervielfältigung oder Verbreitung ohne ausdrückliche Genehmigung von TX Group AG / 20 Minuten ist untersagt.}
\end{quote}

\begin{quote}
    {\textbf{English / Englisch:}} \\[0.3em]
    {\small © 2025. TX Group AG / 20 Minuten.}\\[0.3em]
    {\small This dataset contains copyrighted material from TX Group AG / 20 Minuten. It is provided exclusively for non-commercial scientific research purposes. Any commercial use, reproduction, or distribution without explicit permission from TX Group AG / 20 Minuten is prohibited.}
\end{quote}

\vspace{1cm}

\section{Experiments on Validation Set}
\label{app:thresholds}

\begin{table*}[h!] 
    \renewcommand{\arraystretch}{1.1} 
    \resizebox{\linewidth}{!}{
    \begin{tabular}{@{}lccccc@{}}
    \hline
    \textbf{Model} & \textbf{Above-threshold} & \textbf{Intersection} & \textbf{Union} & \textbf{Best-DE} & \textbf{Best-FR} \\
    \hline
    \texttt{paraphrase-multilingual-mpnet-base-v2} & 61.5 & 46.0 & 61.5 & 47.0 & 46.0 \\
    \texttt{LaBSE} & 66.0 & 50.5 & 50.5 & 50.5 & 50.5 \\
    \texttt{sentence-swissBERT} & 74.5 & 69.5 & 74.5 & 73.0 & 74.5\\
    \texttt{gte-multilingual-base} & 65.0 & 65.0 & 65.0 & 60.0 & 56.0 \\
    \texttt{gte-modernbert-base} & 66.0 & 66.0 & 66.0 & 66.0 & 63.0 \\
    \hline

    \end{tabular}
    }
    \caption{\label{tab:thresh_comparison} Optimal threshold values for different models and alignment approaches. The corresponding F1 scores are presented in Table~\ref{tab:model_comparison}.}
\end{table*}

\end{document}